\pdfoutput=1
\RequirePackage{shellesc}

\documentclass[11pt]{article}

\usepackage{emnlp2021}
\usepackage{graphicx}
\usepackage{float}
\graphicspath{ {./images/} }

\usepackage{times}
\usepackage{latexsym}
\usepackage[autosize]{dot2texi}
\usepackage{tikz}
\usetikzlibrary{shapes,arrows}

\usepackage[T1]{fontenc}

\usepackage[utf8]{inputenc}

\usepackage{microtype}

\title{LightTag: Text Annotation Platform}

\author{Tal Perry \\
  LightTag  \\
  \texttt{tal@lighttag.io} 
  }

\begin{document}
\maketitle
\begin{abstract}
Text annotation tools assume that their user's goal is to create a labeled corpus. However,  users view annotation as a necessary evil on the way to deliver business value through NLP. Thus an annotation tool should optimize for the throughput of the global NLP process, not only the productivity of individual annotators. LightTag is a text annotation tool designed and built on that principle. This paper shares our design rationale, data modeling choices, and user interface decisions then illustrates how those choices serve the full NLP lifecycle.  

\end{abstract}

\section{Introduction}

Building supervised learning models is like operating a manufacturing plant. Raw materials(data) need to be refined and processed(annotated) as a precursor to final assembly. Some manufacturing plants rely on a supply chain (outsource annotation) while others are vertically integrated (annotate in house).  According to the theory of constraints \citep{goldratt2016goal}, a manufacturing process should optimize the global throughput and not any individual sub-process  . 

LightTag is a text annotation tool built on the premise of global optimization by addressing annotator as well as project managers and data scientists who manage the work and enforce production quality. LightTag is a commercial offering with an unlimited free tier for academic use \footnote{Academic free tier avalible at \url{https://lighttag.io/signup/academic}}. LightTag is unique not only in philosophical outlook but also in it's technical implementation and user interface choices,  which we share in this paper. 

The remainder of this article is structured as follows. Section 2 describes prior art. Section 3 analyzes requirements and user personas to derive LightTag's goal. Section 4 describes novel user facing features. Section 5 highlights LightTag's data model and it's implications. We conclude with a number of case studies from industry and academia.  
\section{Related Work}
Emacs \citep{stallman1981emacs} was (shockingly) used to annotate the Penn Treebank \citep{marcus1993building}. Afterwards a series of standalone annotation tools emerged such as Salsa \citep{erk2003salsa} and ITU \citep{eryiugit2007itu} for treebanks or BOEMIE \citep{fragkou2008boemie} and ABNER \citep{settles2005abner}for the biomedical domain. This generation of tools is notable for being standalone software as opposed to the later web-based tools. DUALIST \citep{settles2011closing} stands out as an influential system due to it's inclusion of active learning and feature labeling. 

The following generation of annotation tools were the first to leverage the browser as a user interface platform and include the Brandeis Annotation Tool \citep{verhagen2010brandeis}, GATE Teamware \citep{cunningham2011text}, BRAT \citep{stenetorp2012brat} and WebAnno \citep{yimam2013webanno}. These also leveraged a client-server architecture to enable multi-user annotation projects and server side automation. 
 The recent trends and ubiquity of NLP, along with improved web development frameworks and simplified delivery mechanisms,   have inspired a new generation of tools which cater to data scientists as opposed to academics and emphasize ergonomics. This generation of tools, of which LightTag is a contemporary, include  the open source Docanno \citep{doccano} as well as the commercial Prodi.gy \citep{Prodigy:2018} which focuses on annotator productivity via active learning, and TagTog \citep{cejuela2014tagtog}  which optimizes for bio-medical annotation. 
 
 LightTag's generation of annotation tools offer roughly the same set of capabilities as the previous generation, that of WebAnno, INCEpTION and BRAT. Yet the current generation of tools enjoys a measure of commercial success, despite established and free alternatives. We posit that the current generation of tools has a stronger focus on user experience, ease of use and integration with the end users goals and systems. Thus, despite the similar feature sets between the two generations, we offer the commercial success of LightTag and it's contemporaries as proof of innovation that satisfies previously unmet needs.

\section{Goals and Design}
In designing LightTag, we relied on the manufacturing metaphor mentioned above and identified three user personas and five broad needs that need to be served to optimize the overall "NLP process" as opposed to the local-maxima of individual annotator.

We assume that the end user's goal is to solve a business problem with NLP and that text annotation is a bottleneck in that process \citep{49953} . We distinguish between the rate at which labeled data is produced, and the rate at which labeled data propagates through the end user's NLP process and optimize for the latter.  

\subsection{Requirements Of An Annotation Tool}

\textbf{Expressivity} 
An annotation tool should allow the user to express the kinds of annotation they need to carry out. LightTag supports span annotations, single and multi-label document classification and relationship annotation, including dependency and constituency grammars. LightTag also emphasizes working with "text in the wild" and supports RTL languages, unicode, and very long documents such as legal contracts and electronic medical records. 

\textbf{Productivity} In our taxonomy, productivity is the rate at which an annotator can express the required annotation. All else being equal, the desired productivity is "As much as possible."

\textbf{Coordination} Larger annotation projects need to coordinate the work among the annotators. This can be as simple as sending out N examples to be labeled by  K annotators such that M annotators annotate each example. More complex requirements include sending out tasks to subsets of annotators (based on language or security clearance) or dynamically scheduling work based on agreement levels. 

\textbf{Review and Quality Control}  As in manufacturing, the quality of an annotation needs to be reviewed before delivery. The ability to efficiently review annotations from multiple annotators and/or models is required for larger annotation projects. 

\textbf{Analytics} Project managers and data consumers need to know what is happening. That can include the project's progress, inter-annotator agreement, or annotator accuracy.

\subsection{User Personas}
Modern annotation projects have multiple, distinct, participants whose requirements from an annotation tool differ.  LightTag recognizes three primary user personas annotators, data scientists, and project managers
\begin{figure}[H]
\fbox{\includegraphics[width=\columnwidth]{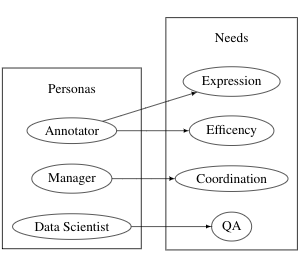}}
\caption{A visualization of the mapping between user personas and their requirements. An annotation platform caters to multiple personas who have different needs.}
\end{figure}
\textbf{Annotators} have three primary needs from an annotation tool. First, they should express the required annotation (an entity, a document class, relationships). Second, the tooling should help annotators avoid errors such as mistakenly annotating trailing whitespace. Third, the annotator's throughput should be maximized subject to their other requirements. 

\textbf{Project Managers} need to control what work is being done and understand the project's cadence and productivity. A common best practice \citep{hovy2010towards} is to have more than one annotator annotate each example. However, coordinating and distributing the work is complex, and the effort scales with the number of annotators while being constrained by the availability of the project manager. LightTag resolves this issue by automating the distribution and management of work according to a project manager's configuration. 

\textbf{Data Scientists} are the final consumer of labeled data and are responsible for assessing it is quality and suitability. LightTag minimizes their heavy lifting by calculating inter-annotator agreement, precision and recall (based on reviewed data), and other metrics. This allows data scientists to spend more time in differentiated data science instead of joining excel files.

\section{User Interface and User Facing Features}
In this section, we present user interface decisions and user-facing features that are, to our knowledge, unique to LightTag. 

\subsection{Annotation Features}

\textbf{Contextual Display:}
Conversational annotation requires preceding messages in order to interpret and properly annotate their followers. LightTag supports this ability through "contextual display," whereby a project manager can configure to display all examples with a particular metadata attribute (such as conversation\_id) at once and sort the items by a separate attribute (such as timestamp).  Thus annotators can see the entire conversation but annotate each message individually.

\textbf{Drag And Drop Relationship Annotation:}
Relationship annotation is a common feature of text annotation tools. To our knowledge, all text annotation tools that offer this functionality implement it as arcs drawn between entities in text, implemented with Scalable Vector Graphics (SVG). 

LightTag implements relationship annotation via the dragging and dropping of entities onto each other and visualizes a full tree in a separate pane. Inspired by the Trees3 program \citet{phillips1998teaching}, users can annotate partial trees and drag and drop branches to annotate richer structures. 

Of note is the ability to annotate constituency grammars by defining non-terminal nodes. This feature is often used to "group together" related nodes in a "container" such as in resume annotation, where a title, company and dates are all constituents of a single job. 
\begin{figure}[H]
\center 

\fbox{\includegraphics[width=\columnwidth]{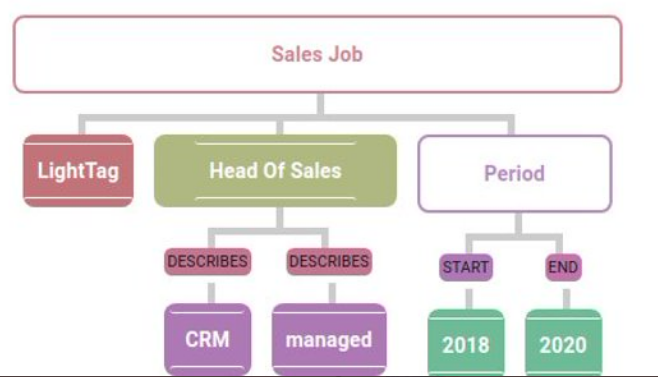}}
\caption{Relationship annotation of a resume with a constituency grammar. The "Sales Job" and "Period" nodes are user defined non-terminals while the other nodes are entities from the text }
\label{fig:rel}
\end{figure}

\textbf{Large Taxonomies:}
Annotation starts with a taxonomy, the collection of concepts that will be annotated. Some projects are based on taxonomies with hundreds or thousands of classes or entity types. In these cases, it is infeasible to display the entire taxonomy in a static list. Long lists slow down annotators and introduces an availability bias \citep{tversky1973availability} where annotators are more likely to select entities that are visible and at the top of a list, thus biasing the resulting data.

LightTag resolves this issue by providing a searchable field for classes and entities, allowing the annotator to quickly find the correct class by searching. 
\begin{figure}[h]

\fbox{\includegraphics[width=\columnwidth]{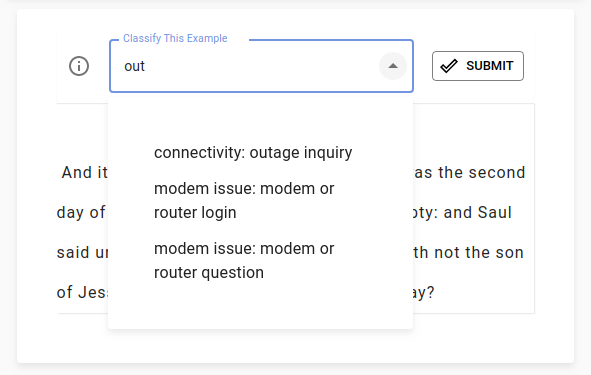}}
\caption{The user can search a taxonomy of a few thousand classes to quickly find the most relevant class, without scrolling through a list}

\end{figure}

\textbf{Unobtrusive Pre-Annotations:}
Many annotation tools offer pre-annotations to increase annotator productivity. The efficacy of pre-annotations depends on both their accuracy and how the user interacts with them, particularly when the pre-annotations are incorrect. If a user must make an action for every pre-annotation, incorrect ones risk increasing the total number of actions and diminishing productivity. 
\begin{figure}[H]
\center
\fbox{\includegraphics[width=0.9\columnwidth]{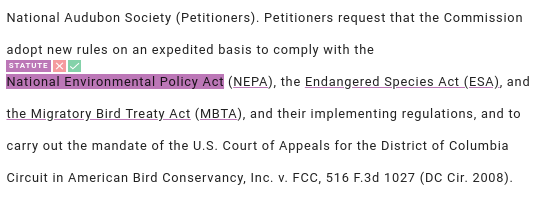}}
\caption{Unobstrusive pre-annotations are displayed as colored underlines. When the user hovers over a pre-annotation they can accept or reject it. A batch accept butto (not displayed) allows users to save clicks by accepting all at once.  }
\end{figure}

LightTag displays pre-annotations in as an unobtrusive underline. The user can ignore them (and thus take no action) or accept/reject them by hovering over a pre-annotation and clicking. LightTag offers a batch accept button allowing users to accept many pre-annotations at once. 

We find that this mode of interaction has a significant effect on annotator productivity, with a near doubling of annotator throughput achieved when only 50\% of pre-annotations are accepted.

\textbf{Annotating With Search}
Like other annotation tools, LightTag defaults to displaying examples to annotate one at a time. However, many datasets are sparse with respect to the classes or entities that users need to annotate. In such cases having annotators annotate each example, where the majority are irrelevant, is ineffective. 

To address this issue, LightTag follows \citet{Attenberg:2010} by offering a "Search Mode" in which the entire dataset is displayed in an infinite scroll, and the user can narrow it down using search queries. 

LightTag's  implementation of search is noteworthy because it is operationally simple while remaining fast at scale.    \citet{cox2012regular} demonstrated the use of tri-gram indices to speed up plain text and regular expression search and \citet{korotkov2012index} introduced an implementation to Postgres. Leveraging these, LightTag can offer users very fast regular expressions search with minimal operational overhead. 

\begin{figure}[H]

\fbox{\includegraphics[width=\columnwidth]{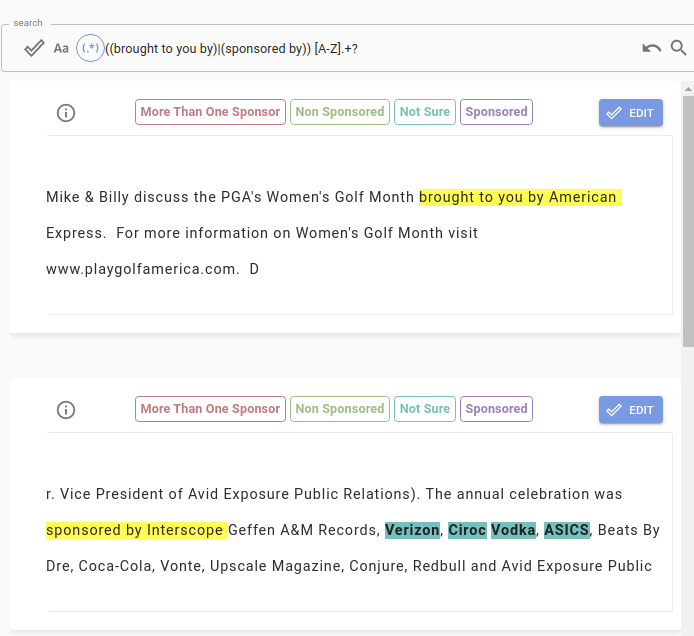}}
\caption{Annotating with search. Users can write search queries or regular expressions to narrow down the set of documents to work on. In this example, documents from the Federal Registar are annoted for mentions of foreign policy. }

\end{figure}

\subsection{Review}
Project managers and data scientists want to review annotations produced by both annotators and, later, by models. LightTag's Review mode displays all annotations made in a selected example and consolidates agreements and conflicts. Reviewers can narrow the scope of review to human or model annotations and automatically accept all annotations that meet a certain agreement threshold. 
\begin{figure}[H]
\fbox{\includegraphics[width=\columnwidth]{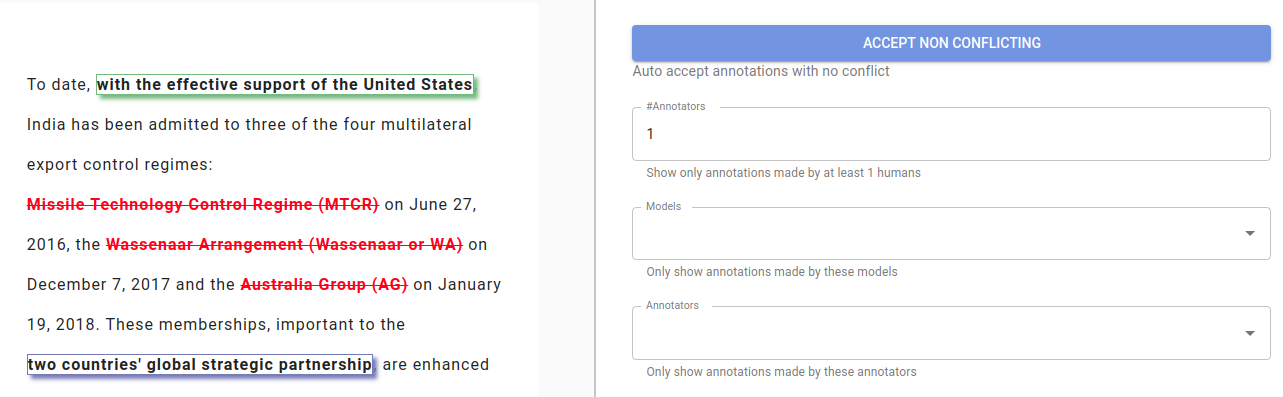}}
\caption{Agreement detection powered by the relational model. Conflicts are easily detected by the system and visually displayed during review. A reviewer can click on the button to accept all annotations meeting a specific criteria}

\end{figure}

\subsubsection{Batch Lexical Review}
We observe that the distribution of annotated entities is Zipfian. Rather than having reviewers review every case of trivially correct or incorrect annotations, LightTag offers a batch review function where every instance of a particular lexical form can be seen and reviewed in either a stream or in one click.
\begin{figure}[H]
\center
\fbox{\includegraphics[width=\columnwidth]{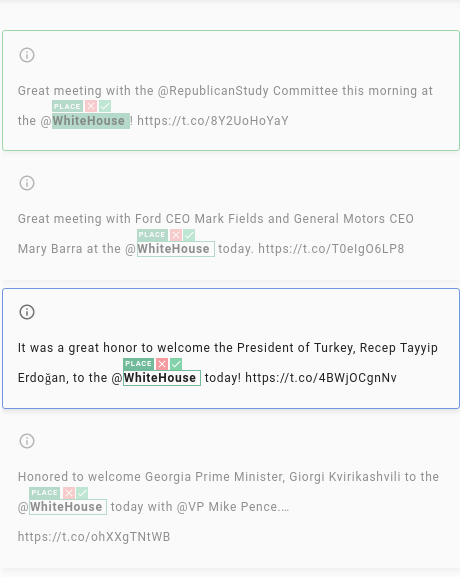}}
\caption{All instances of the form "White House" labeled as place are displayed. The user can review them one by one or batch accept/reject them with one click.}

\end{figure}

\section{Backend and Data Model}
LightTag’s focus on project management and quality assurance requires a rich data management structure. LightTag’s backend is a relational database using Postgres and makes heavy use of relational design theory \citet{codd2002relational}. In this section, we provide an overview of LightTag's data model and elaborate on useful implications.

\subsection{Relational Data Model}
A project manager in LightTag may define a Job comprised of the Dataset to annotate and the concepts (entity tags or document classes) with which to annotate. N annotators should annotate each Example in the Dataset of a Job. A project manager may wish to have the same Dataset annotated with the same Schema in two Jobs, where a different Team executes each Job.  The definition and assignment of work as described above fits neatly into a relational model. 
 \begin{figure}
\includegraphics[width=\columnwidth]{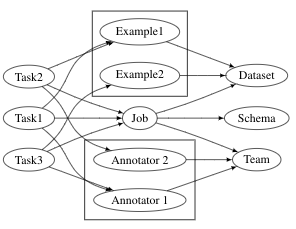}
\caption{A graphical display of the relation between data entities describing three tasks carried out by two annotators as part of a job }
\end{figure}

The natural extension of a relational data model is that annotations are stored separately from the Example being annotated. LightTag takes this idea a step further and separates the Platonic Ideal \citep{plato1961republic} of annotation from the event that Annotator A made Annotation X, thus brining the database to third normal form. 
For example, the "Ideal" that  “Document X is classified as class Y” is stored in a distinct table with id Z. A separate events table would then store the event “Annotator 1 made classification Z during Task x”. Storing every possible ideal would be inefficient, thus LightTag stores the ideal of an annotation the first time it is manifested via annotation.

\begin{figure}[H]
\includegraphics[width=\columnwidth]{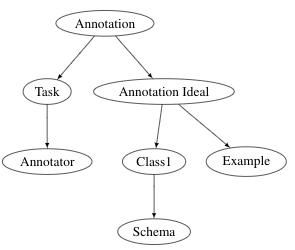}
\caption{A graphical depiction of the relations defining an annotation. Each node corresponds to a row in the respective table.  The annotator that worked on the Task made the annotation represent by the Ideal of an annotation that the Example is classified as Class1}
\end{figure}

\subsection{Relational Data Implications}
A notable implication of this design is batch functionality during review. For example, automatically accepting all annotations with a majority vote is displayed as a button to the user and is implemented by aggregating over the “Annotation Ideal table” id, counting and comparing with the number of users that saw that example (derived from the Tasks table). 

\textbf{Measuring Negative Annotations}
When annotating with a larger team, we can not assume that every team member annotated every example. Thus when calculating metrics such as inter-annotator agreement, a particular annotator even saw the particular example needs to be accounted for. The relational model makes this easy by implicitly providing a list for each annotator of the Examples they worked on (by aggregating on the Task table). 

\textbf{Majority Vote}
During a quality assurance process, it is common practice to automatically accept annotations with a majority or unanimous vote automatically and manually review annotations in a conflicting state. By separating the Ideal of an Annotation from the Event that annotation was made and recording the particular Job under which the annotation was made, LightTag can provide the reviewing user with a one-click functionality to accept all annotations that meet some agreement criteria.  

\textbf{Transitive Annotation Rejections}
LightTag's quality assurance functionality assumes only one correct answer for an annotated span or a document classification\footnote{In single-class classification. In configurations where more than one class is allowed per document, this assumption is removed}. When a reviewer marks an annotation as correct, the system rejects any conflicting annotation automatically, be it a difference in class, an entity tag, or span range. If annotations A and B overlap and A is correct, then B must be incorrect. The relational model allows executing the transitive rejection in $\mathcal{O}(1)$ time instead of scanning the entire annotation table. More importantly, doing so in a single database transaction ensures that the data is never in an invalid state.

\section{Case Studies}
\subsection{Detecting Foreign Policy With Search}
The Federal Register is the official journal of the federal government of the United States that contains government agency rules, proposed rules, and public notices. A team of researchers from Harvard Law wished to annotate every mention of foreign policy across over 100,000 rules spanning 2.1 Million paragraphs. 
A team of 15 undergraduate law students was assembled, and the data was loaded into LightTag. Using LightTag's search mode, subsections of the dataset were assigned to subsets of annotators who then searched over the corpus to find and annotate over 60 thousand distinct mentions of foreign policy in the corpus. 

\subsection{Sponsorship Detection in Podcasts}
Thoughtleaders (TL). a provider of marketing analytics created a corpus of podcast transcripts to detect which brands sponsored each podcast episode \citep{kassuto_2021}. TL trained a BERT-based model to recognize brands and distinguish between casual brand mentions and mentions of a podcast sponsor. 
To create a training corpus with LightTag, TL first created pre-annotations with regular expressions and then had their team validate those and annotate missing entities. 

Within a week, they had generated over 20 thousand human-annotated entities and trained a model that met their requirements. To validate the model's performance, they loaded model predictions from data outside of the training set into LightTag and used the review functionality to verify model predictions and establish performance metrics manually. 
\subsection{Multi-Lingual Malware Detection}
CS is a provider of Malware analytic and early detection systems. To serve their customers, they develop custom NLP models to detect the sale of zero-day exploits on the dark web. Due to the multi-lingual nature of the data, they needed to manage multiple teams and projects, each specializing in a particular language (Mandarin, Russian, English, etc.). LightTag's workforce management solution enabled them to minimize project management overhead, while pre-annotations and review functionality allowed the team to validate both annotations and candidate model outputs, reaching production grade models and their market faster.  

\subsection{Mentions In Other Publications}

\citet{sarkardetecting} created a corpus for emotion detection in musical lyrics. \citet{vasilyev2020sensitivity} generated a corpus of text-summary quality on a five-point scale across five attributes of the summary.\citet{alnazzawi2021building} annotated a joint corpus of tweets and electronic health records to detect underlying risk factors for hypertension and diabetes.\citet{pitenis2020offensive} developed a Greek language corpus of offensive language

\bibliography{anthology,custom}

\begin{thebibliography}{29}
\expandafter\ifx\csname natexlab\endcsname\relax\def\natexlab#1{#1}\fi

\bibitem[{Alnazzawi(2021)}]{alnazzawi2021building}
Noha Alnazzawi. 2021.
\newblock Building a semantically annotated corpus for chronic disease
  complications using two document types.
\newblock \emph{PloS one}, 16(3):e0247319.

\bibitem[{Attenberg and Provost(2010)}]{Attenberg:2010}
Josh Attenberg and Foster Provost. 2010.
\newblock \href {https://doi.org/10.1145/1835804.1835859} {Why label when you
  can search? alternatives to active learning for applying human resources to
  build classification models under extreme class imbalance}.
\newblock In \emph{Proceedings of the 16th ACM SIGKDD International Conference
  on Knowledge Discovery and Data Mining}, KDD '10, page 423–432, New York,
  NY, USA. Association for Computing Machinery.

\bibitem[{Cejuela et~al.(2014)Cejuela, McQuilton, Ponting, Marygold,
  Stefancsik, Millburn, Rost, Consortium et~al.}]{cejuela2014tagtog}
Juan~Miguel Cejuela, Peter McQuilton, Laura Ponting, Steven~J Marygold, Raymund
  Stefancsik, Gillian~H Millburn, Burkhard Rost, FlyBase Consortium, et~al.
  2014.
\newblock tagtog: interactive and text-mining-assisted annotation of gene
  mentions in plos full-text articles.
\newblock \emph{Database}, 2014.

\bibitem[{Codd(2002)}]{codd2002relational}
Edgar~F Codd. 2002.
\newblock A relational model of data for large shared data banks.
\newblock In \emph{Software pioneers}, pages 263--294. Springer.

\bibitem[{Cox(2012)}]{cox2012regular}
Russ Cox. 2012.
\newblock Regular expression matching with a trigram index or how google code
  search worked.

\bibitem[{Cunningham and Bontcheva(2011)}]{cunningham2011text}
Diana Maynard~Hamish Cunningham and Kalina Bontcheva. 2011.
\newblock \emph{Text Processing with GATE (Version 6).}
\newblock University of Sheffield D.

\bibitem[{Erk et~al.(2003)Erk, Kowalski, and Pado}]{erk2003salsa}
Katrin Erk, Andrea Kowalski, and Sebastian Pado. 2003.
\newblock The salsa annotation tool.
\newblock In \emph{Proceedings of the Workshop on Prospects and Advances in the
  Syntax/Semantics Interface}, pages 1--4.

\bibitem[{Eryi{\u{g}}it(2007)}]{eryiugit2007itu}
G{\"u}l{\c{s}}en Eryi{\u{g}}it. 2007.
\newblock Itu treebank annotation tool.
\newblock In \emph{Proceedings of the Linguistic Annotation Workshop}, pages
  117--120.

\bibitem[{Fragkou et~al.(2008)Fragkou, Petasis, Theodorakos, Karkaletsis, and
  Spyropoulos}]{fragkou2008boemie}
Pavlina Fragkou, Georgios Petasis, Aris Theodorakos, Vangelis Karkaletsis, and
  Constantine~D Spyropoulos. 2008.
\newblock Boemie ontology-based text annotation tool.
\newblock In \emph{LREC}. Citeseer.

\bibitem[{Goldratt and Cox(2016)}]{goldratt2016goal}
Eliyahu~M Goldratt and Jeff Cox. 2016.
\newblock \emph{The goal: a process of ongoing improvement}.
\newblock Routledge.

\bibitem[{Hovy and Lavid(2010)}]{hovy2010towards}
Eduard Hovy and Julia Lavid. 2010.
\newblock Towards a ‘science’of corpus annotation: a new methodological
  challenge for corpus linguistics.
\newblock \emph{International journal of translation}, 22(1):13--36.

\bibitem[{Kassuto(2021)}]{kassuto_2021}
Avi Kassuto. 2021.
\newblock \href {https://www.thoughtleaders.io/blog/read-between-lines} {Read
  between the lines}.

\bibitem[{Korotkov(2012)}]{korotkov2012index}
Alexander Korotkov. 2012.
\newblock Index support for regular expression search.
\newblock In \emph{Proc. PostgreSQL Conference}.

\bibitem[{Marcus et~al.(1993)Marcus, Santorini, and
  Marcinkiewicz}]{marcus1993building}
Mitchell Marcus, Beatrice Santorini, and Mary~Ann Marcinkiewicz. 1993.
\newblock Building a large annotated corpus of english: The penn treebank.

\bibitem[{Montani and Honnibal(2018)}]{Prodigy:2018}
Ines Montani and Matthew Honnibal. 2018.
\newblock \href {http://arxiv.org/abs/to appear} {Prodigy: A new annotation
  tool for radically efficient machine teaching}.
\newblock \emph{Artificial Intelligence}, to appear.

\bibitem[{Nakayama et~al.(2018)Nakayama, Kubo, Kamura, Taniguchi, and
  Liang}]{doccano}
Hiroki Nakayama, Takahiro Kubo, Junya Kamura, Yasufumi Taniguchi, and Xu~Liang.
  2018.
\newblock \href {https://github.com/doccano/doccano} {{doccano}: Text
  annotation tool for human}.
\newblock Software available from https://github.com/doccano/doccano.

\bibitem[{Phillips(1998)}]{phillips1998teaching}
Colin Phillips. 1998.
\newblock Teaching syntax with trees.
\newblock \emph{Glot international}, 3.

\bibitem[{Pitenis et~al.(2020)Pitenis, Zampieri, and
  Ranasinghe}]{pitenis2020offensive}
Zeses Pitenis, Marcos Zampieri, and Tharindu Ranasinghe. 2020.
\newblock Offensive language identification in greek.
\newblock \emph{arXiv preprint arXiv:2003.07459}.

\bibitem[{Plato(1961)}]{plato1961republic}
HG~Plato. 1961.
\newblock \emph{Republic}.
\newblock Princeton University Press.

\bibitem[{Sambasivan et~al.(2021)Sambasivan, Kapania, Highfill, Akrong,
  Paritosh, and Aroyo}]{49953}
Nithya Sambasivan, Shivani Kapania, Hannah Highfill, Diana Akrong,
  Praveen~Kumar Paritosh, and Lora~Mois Aroyo. 2021.
\newblock "everyone wants to do the model work, not the data work": Data
  cascades in high-stakes ai.

\bibitem[{Sarkar(2020)}]{sarkardetecting}
Diptanu Sarkar. 2020.
\newblock Detecting emotions in lyrics.

\bibitem[{Settles(2005)}]{settles2005abner}
Burr Settles. 2005.
\newblock Abner: an open source tool for automatically tagging genes, proteins
  and other entity names in text.
\newblock \emph{Bioinformatics}, 21(14):3191--3192.

\bibitem[{Settles(2011)}]{settles2011closing}
Burr Settles. 2011.
\newblock Closing the loop: Fast, interactive semi-supervised annotation with
  queries on features and instances.
\newblock In \emph{Proceedings of the 2011 Conference on Empirical Methods in
  Natural Language Processing}, pages 1467--1478.

\bibitem[{Stallman(1981)}]{stallman1981emacs}
Richard~M Stallman. 1981.
\newblock Emacs the extensible, customizable self-documenting display editor.
\newblock In \emph{Proceedings of the ACM SIGPLAN SIGOA symposium on Text
  manipulation}, pages 147--156.

\bibitem[{Stenetorp et~al.(2012)Stenetorp, Pyysalo, Topi{\'c}, Ohta, Ananiadou,
  and Tsujii}]{stenetorp2012brat}
Pontus Stenetorp, Sampo Pyysalo, Goran Topi{\'c}, Tomoko Ohta, Sophia
  Ananiadou, and Jun’ichi Tsujii. 2012.
\newblock Brat: a web-based tool for nlp-assisted text annotation.
\newblock In \emph{Proceedings of the Demonstrations at the 13th Conference of
  the European Chapter of the Association for Computational Linguistics}, pages
  102--107.

\bibitem[{Tversky and Kahneman(1973)}]{tversky1973availability}
Amos Tversky and Daniel Kahneman. 1973.
\newblock Availability: A heuristic for judging frequency and probability.
\newblock \emph{Cognitive psychology}, 5(2):207--232.

\bibitem[{Vasilyev et~al.(2020)Vasilyev, Dharnidharka, Egan, Chambliss, and
  Bohannon}]{vasilyev2020sensitivity}
Oleg Vasilyev, Vedant Dharnidharka, Nicholas Egan, Charlene Chambliss, and John
  Bohannon. 2020.
\newblock Sensitivity of blanc to human-scored qualities of text summaries.
\newblock \emph{arXiv preprint arXiv:2010.06716}.

\bibitem[{Verhagen(2010)}]{verhagen2010brandeis}
Marc Verhagen. 2010.
\newblock The brandeis annotation tool.
\newblock In \emph{LREC}.

\bibitem[{Yimam et~al.(2013)Yimam, Gurevych, de~Castilho, and
  Biemann}]{yimam2013webanno}
Seid~Muhie Yimam, Iryna Gurevych, Richard~Eckart de~Castilho, and Chris
  Biemann. 2013.
\newblock Webanno: A flexible, web-based and visually supported system for
  distributed annotations.
\newblock In \emph{Proceedings of the 51st Annual Meeting of the Association
  for Computational Linguistics: System Demonstrations}, pages 1--6.

\end{thebibliography}
\bibliographystyle{acl_natbib}

\end{document}